\documentclass[12pt]{article}
\usepackage{graphicx}
\usepackage{latexsym}

\usepackage[utf8]{inputenc}
\usepackage{url}
\usepackage{amssymb,amsmath}
 \newcommand{\Bem}[1]{}

\newcommand{\figaddr}[1]{#1}

\begin{document}

\title{Explainable Graph Spectral Clustering of Text Documents}
 
\author{Bart{\l}omiej Starosta, Mieczys{\l}aw A. K{\l}opotek,\\ S{\l}awomir T. Wierzcho{\'n}
\\
Institute of Computer Science,  \\Polish Academy of Sciences, \\Warsaw, Poland}

\date{}

\maketitle

\begin{abstract}  
Spectral clustering methods are known for their ability to represent clusters of diverse shapes, densities etc. 
However,  results of such algorithms, when applied e.g. to text documents, are  hard to explain to the user, especially due to embedding in the spectral space which has no obvious relation to document contents.  Therefore there is an urgent need to elaborate methods for explaining the outcome of the clustering.  This paper presents a contribution towards this goal. We present a proposal of explanation of results of combinatorial Laplacian based graph spectral clustering.
It is based on showing (approximate)  equivalence of combinatorial Laplacian embedding,  $K$-embedding (proposed in this paper) and term vector space embedding. Hence a bridge is constructed between the textual contents and the clustering results. We provide theoretical background for this approach. We performed experimental study showing that $K$-embedding approximates well Laplacian embedding under favourable block matrix conditions and show that approximation is good enough under other conditions. 
\end{abstract}

\section{Introduction}

The focus of our research is on the so-called graph spectral cluster analysis (GSC) in application to sparse {data}sets. Spectral clustering methods  are known for their ability to represent clusters of diverse shapes, densities, etc. They constitute an approximation to graph cuts of various types (plain cuts, normalized cuts, ratio cuts). They are applicable to unweighted and  weighted similarity graphs.  In spite of their advantages, various shortcomings were encountered for their application.
One of them is the explainability problem. 
Typically, it is recommended 
\footnote{
see e.g. 
\url{https://crunchingthedata.com/when-to-use-spectral-clustering/}
}
\emph{not to use } spectral clustering if you 
"need an explainable algorithm".
\Bem{
, like dimensionality curse (for both space and processing time),  sensitivity to noise, number-of-clusters issues, availability of multiple representations of the same similarity relations (e.g. in text clustering: tfidf representation, tf representation, various brands of transformer related {embedding}s etc.),  variety of possible {embedding}s in Euclidean space as well as kernel {embedding}s. These shortcomings urged the development of dozens of variants of clustering algorithms which may yield distinct clusterings especially under not clear cut settings. 
This means in essence that 
}
A result of clustering is hard to explain to the user solely on the grounds of numerical results of the clustering algorithm, especially due to their embedding in the spectral space.
Therefore there is an urgent need to elaborate methods for explaining the outcome of the spectral clustering, as already described. 
\Bem{But in order to pursue the explanation task, the properties of the spectral clustering algorithms themselves need to be explored in more depth.} 
This is of practical importance as SGC is used frequently in the context of natural language processing (e.g. \cite{JANANI2019192}, \cite{10.1007/978-3-642-28601-8_15},\cite{DBLP:journals/corr/SedocGUF16}, \cite{cui2023free}).

 This paper presents a contribution towards this goal. We propose an  explanation method of results of combinatorial Laplacian based graph spectral clustering.
It is based on showing (approximate)  equivalence of combinatorial Laplacian embedding, our $K$-embedding and term vector space embedding.
But if we use the term vector space embedding, we can explain cluster membership of textual documents by pointing at significant words/terms, as   commonly practiced \cite{Nareddy:2011}, subject to various improvements \cite{Role:2014}, \cite{Poostchi:2018} (compare also with Shapley-value based approaches \cite{merrick:2020explanation}).  
Section \ref{sec:prevResearch} reviews previous research on clustering explanations. 
Section \ref{sec:ourTheory} introduces our proposal for explanation of combinatorial Laplacian based spectral clustering. Section \ref{sec:experiments} presents some experimental results on clustering using the combinatorial Laplacian based clustering and our $K$-embedding based clustering. 
We conclude the paper with some final remarks in Section \ref{sec:conclusions}.

\section{Previous Research}\label{sec:prevResearch}

As the realm of clustering algorithms is vast, see e.g.  \cite{EZUGWU2022104743}, or \cite{STWMAKSpringer:2018}, 
 we narrow our interest to the large family of spectral clustering algorithms \cite{vonLuxburg:2007}, \cite{pmlr-v162-macgregor22a}, \cite{Xu2021}, 
 which have numerous desirable properties (like detection of clusters with various shapes, applicability to high dimensional {data}sets, capability to handle categorical variables), yet they suffer from various shortcomings, common to other sets of algorithms, including multiple possibilities of representation of the same dataset, producing results in a space different from the space of original problem, curse of dimensionality etc.  which are particularly grieving under large and sparse data set scenario.   

Let us briefly recall the typical spectral clustering algorithm in order to make it understandable, how distant the clustering may be from the applier's comprehension \cite{vonLuxburg:2007}. 
The first step consists in creating a similarity matrix of objects (in case of documents based on tf, tfidf, in unigram or n-gram versions, or some transformer based {embedding}s are the options -- consult e.g. \cite{Man08} for details), then mixing them in case of multiple views available.\Bem{, then computation of a Laplacian (at least three brands are used: combinatorial, normalized, random-walk \cite{vonLuxburg:2007}, or some kernel-based versions, other options non-backtracking matrix \cite{Krzakala:2013}, degree-corrected versions of the modularity matrix \cite{TIOMOKO:2018} or the Bethe-Hessian matrix \cite{arxiv1406_1880}
),} The second step is to calculate a Laplacian matrix. There are at least three variants to use: combinatorial, normalized, and random-walk Laplacian, \cite{vonLuxburg:2007}. Other options are also possible, like: some kernel-based versions, non-backtracking matrix \cite{Krzakala:2013}, degree-corrected versions of the modularity matrix \cite{TIOMOKO:2018} or the Bethe-Hessian matrix  \cite{arxiv1406_1880}. Then computing eigenvectors and eigenvalues, eigenvector smoothing (to remove noise and/or achieve robustness against outliers) choice of eigenvectors, and finally clustering in the space of selected eigenvectors (via e.g. $k$-means). 
The procedure may be more complex, e.g. one may add loops back to preceding steps based on feedback from quality analysis, like degree of deviation from block-structure of the Laplacian.

Cluster Analysis, like the entire domain of Artificial Intelligence, experienced a rapid development over the recent years, providing with algorithms of growing complexity and efficiency that are regrettably characterized by their ``black-box nature'' that is their results are hard to understand by human users and therefore there exists a growing resistance for their application in practical settings. This phenomenon led to development of a branch of AI called ``Explainable Artificial Intelligence'' (XAI) \cite{BARREDOARRIETA202082}, 
with subbranches including Explainable Clustering \cite{arxiv.2112.06580}. 

The ``black box'' problem relates in particular to cluster analysis \cite{arxiv.2112.06580}. \Bem{Here the situation is so much more difficult than in the applications of types classification that the very essence of the concept of ``cluster''  is not well defined.} The situation is more difficult here, compared e.g. to the classification tasks, because the very essence of the concept of ``cluster'' is not well defined.
This is in spite of the fact that the scientific research area of cluster analysis, has nearly a century long history. Hundreds of clustering algorithms have been developed and countless applications are reported. 

\Bem{
Yet its foundations still remain informal. A number of efforts to create an axiomatic system for cluster analysis, like those of Kleinberg \cite{Kleinberg:2002} and other, but no commonly agreed definition of cluster or clustering algorithm was reached though many cluster quality indices have been developed. 
What is more, one of the most frequently cited  axiomatic system of Kleinberg \cite{Kleinberg:2002}, which was claimed to consist of natural cluster properties,  
was controversial from the very beginning. Kleinberg proved himself that the entire axiomatic system is contradictory, and only pairs of these axioms are not contradictory. The axiom of consistency turned out to be controversial as it excludes the $k$-means algorithm from the family of clustering functions. It has been shown in \cite{MAKRAK:2020:fixdimcons}  that consistency is contradictory by itself if we consider fixed dimensional Euclidean spaces which are natural domain of $k$-means application. Other problems are described in \cite{MAK:2017:kleinbergaxioms}.
It seems to be disastrous for the domain of clustering algorithms if an axiomatic system consisting of ``natural axioms'' is self-contradictory. It renders the entire domain questionable. Hence numerous efforts have been made to cure such a situation by proposing different axiom sets or modifying Kleinberg's one.

In particular it is disastrous when one wants to explain the clustering obtained as that the cluster definition is usually determined ad hoc with respect to a specific business application.
}

Though there exist some approaches for cluster explanations like \cite{Kauffmann_2022} that are applied to the outcome of spectral clustering, they are not based on the actual principles of spectral clustering, but rather on cluster approximation with some other algorithms. There exist explanation methods  to components of spectral clustering, that is to $k$-means \cite{arxiv.2111.03193} but they are insufficient to explaining the outcome of spectral clustering. 

What makes the explanation of the clustering results more difficult is the fact   that the selected embedding may have impact 
on clustering results. 
In \cite{STW:MAK:2020:clustermaps}, we have investigated already the impact of Laplacian type embedding on the clustering of regular graphs. It has been demonstrated  that various Laplacians are sensitive to choice of number of clusters  to different degrees, which should be correlated with the closeness of eigenvalues, and also  to methods of similarity computation. However, the effect is not the same for various real {data}sets as well as for Laplacian types and an explanation of such a behaviour is necessary. 

Still another aspect to consider is that, 
under practical settings, the data can be described from different viewpoints.
One has to face the challenge of aggregating such sets or of the corresponding similarity matrices. This challenge became a hot topic in recent time and numerous works have been devoted to this issue, e.g. \cite{Xu13}, ~\cite{Zhao17}. 
The aggregation usually involves uncontrolled weighting schemes which add an additional dimension in the process of clustering explanation, in particular in the light of the theorem of Watanabe \cite{MAK0:1991,MSE1:MAK:2012}.

\section{A Brief Overvierw of  Graph Spectral Clustering}

Spectral clustering methods can be viewed as a relaxation of cut based graph clustering methods. 
Let $S$ be a similarity matrix between pairs of items (e.g. documents). It induces a graph whose nodes correspond to the items.
It is generally assumed that this graph is deemed as one without self-loops (in spite of the fact that an item is most similar to itself). Hence the diagonal of $S$ is assumed to be filled with zeros (see e.g. \cite{TU:2022:3673} as one example of many). 

A(n unnormalised or) combinatorial Laplacian $L$ corresponding to this matrix is defined as 
\begin{equation}\label{eq:combLapDef} L=D-S, \end{equation}
where $D$ is the diagonal matrix with $d_{jj}=\sum_{k=1}^ns_{jk}$ for each $j \in [n]$. 
A normalized Laplacian $\mathcal{L}$ of the graph represented by~$S$ is defined as 
\begin{equation}\label{eq:normLapDef}\mathcal{L}=D^{-1/2}L D^{-1/2}= I -D^{-1/2}S D^{-1/2} .\end{equation}%

Their relationship to cut based clustering is as follows:
The RCut criterion corresponds to finding the partition matrix $P_{RCut} \in \mathbb{R}^{n \times k}$ that minimizes the formula $H' L H$ over the set of all partition matrices $H \in \mathbb{R}^{n\times k}$. Such formulated problem is NP-hard. That is why we relax it by assuming that $H$ is a column orthogonal matrix. In this case the solution is obvious: the columns of $P_{RCut}$ are eigenvectors of $L$ corresponding to $k$ smallest eigenvalues of~$L$. 
Similarly, the columns of matrix $P_{NCut}$, representing NCut criterion, are eigenvectors of $\mathcal{L}$ corresponding to $k$ smallest eigenvalues of~$\mathcal{L}$. For an explanation and further details see e.g.~\cite{vonLuxburg:2007} or~\cite{STWMAKSpringer:2018}.

The RCut clustering can be imagined in more detail as follows: 
 Let $\mathbf{y_j}$ be the vector of indicators of membership in cluster $j$. Indicators will have the form: if the element $i$ belongs to cluster $j$ of cardinality $n_j$, then $y_{ij}=\sqrt{1/n_j}$, and otherwise $y_{ij}=0$.
Then 
$\mathbf{y_j}^TL\mathbf{y_j}=\frac12 \sum_i \sum_l s_{il}(y_{ij}-y_{lj})^2$ 
$=
\frac12  \sum_{i\in C_j} \sum_{l\not\in C_j}  s_{il}
(\sqrt{1/n_j}-0)^2
+\frac12  \sum_{i\not\in C_j} \sum_{l\in C_j}  s_{il}
(0-\sqrt{1/n_j})^2
$
$=\frac{1}{n_j}  \sum_{i\in C_j} \sum_{l\not\in C_j}  s_{il}$.
By summing $\sum_j \mathbf{y_j}^TL\mathbf{y_j}$
we obtain  RCut. All $\mathbf{y_j}$ are orthogonal   and $\|y_j\|=1$. 
So minimizing RCut is minimizing $\sum_j \mathbf{y_j}^TL\mathbf{y_j}$.

Finding $y_i$ would be a hard problem, so we relax $y_i$ component values to be real values $y'_j$ and we minimize  $\sum_j \mathbf{y'_j}^TL\mathbf{y'_j}$ subject to constraints  that  $y'_j$  are to be orthogonal and all $\|y'_j\|^2=n$.
By the Rayleigh-Ritz theorem,   the solution of this problem is given by the vectors $\mathbf{y'_j}$ which are the eigenvectors corresponding to
the $k$ smallest eigenvalues of $L$. 
The minimization is performed by $k$-means in the embedding   $\mathbf{\hat{x}}_i=({y'}_{i1},...{,y'}_{ik})^T$. This embedding  will be called 
 afterwards \emph{$L$-embedding}.
 Note that if the eigenvectors of $L$ were really the $y$ indicator vectors, then $k$-means would achieve the absolute minimum equal zero and return the intrinsic clustering and RCut optimum will be reached. 
 \Bem{
If we would use this embedding, all the clusters were of the same size, and $k$-means would be applied, and the obtained clustering would be true one, then the value of the optimum would have a very interesting form.
}

The disadvantage of \mbox{$L$-embedding} $\mathbf{\hat{x}}_i$ is that there is no direct link between its components ${y'}_{ij}$ ($j=\{1,\dots,k\}.$) and the cosine similarity computation between the elements (textual documents). Hence, a translation to cosine similarity to cluster center is not straightforward. 
In the subsequent subsection we seek a way out of this embarrassing situation.   

\section{Searching For Clustering Explanation} \label{sec:ourTheory}

\subsection{A Proposal of Double-centered  Document Similarity Matrix Based Embedding}

Let us think for a moment about a particular embedding of the nodes of the graph, based on \cite{RAKMAKSTW:2020:trick}. 
Let $A$ be a matrix of the form:
\begin{equation}
    A= \mathbf{1}\mathbf{1}^T-I-S\;,
\end{equation}
\noindent where $I$ is the identity matrix, and $\mathbf{1}$ is the (column) vector consisting of ones, both of appropriate dimensions. 
Note that here we have to assume that the diagonal of $S$ consists of zeros.
Let $K$ be the matrix of the (double centered) form \cite{Gower:1966}:
\begin{equation}
    K=-\frac12(I-\frac1{n}\mathbf{1}\mathbf{1}^T)A(I-\frac1{n}\mathbf{1}\mathbf{1}^T)\;,
\end{equation}
\noindent with $n\times n$ being the dimension of $S$. 
Note that $\mathbf{1}$ is an eigenvector of $K$, with the corresponding eigenvalue equal to 0. 
In fact, $(I-\frac1{n}\mathbf{1}\mathbf{1}^T)\mathbf{1}
=\mathbf{1}-\frac1{n}\mathbf{1}\mathbf{1}^T\mathbf{1}
=\mathbf{1}-\frac1{n}\mathbf{1}n=\mathbf{0}
$. All the other eigenvectors must be orthogonal to it as $K$ is real and symmetric, so for any other eigenvector $\mathbf{v}$ of $K$ we have: $\mathbf{1}^T\mathbf{v}=0$.

Let $\Lambda$ be the diagonal matrix of eigenvalues of $K$, and $V$ the matrix where columns are corresponding (unit length) eigenvectors of $K$. 
Then $K=V\Lambda V^T$. 
Let $\mathbf{z}_i=\Lambda^{1/2} V^T_{i}$, where $V_i$ stands for $i$-th row of $V$.  
Let $\mathbf{z}_i,\mathbf{z}_\ell$ be the {embedding}s of the nodes $i,\ell$, resp. 
This embedding shall be called \emph{$K$-embedding}. 
Then 
\begin{equation}    \label{eq:embeddist}
\|\mathbf{z}_i-\mathbf{z}_\ell\|^2= 1-S_{i\ell}
\end{equation}
for $i\ne \ell$ \Bem{Bylo: $i \ne j$}. Hence upon performing $k$-means clustering in this space we \emph{de facto} try to maximize the sum of similarities within a cluster. \footnote{
Lingoes correction is needed, if $K$ turns out to have negative eigenvalues, see \cite{RAKMAKSTW:2020:trick}. 
The correction consists in adding $2\sigma$ to all elements of dissimilarity matrix $A$ except for the main diagonal, which has to stay equal to zero, where $\sigma\ge -\lambda_m$  where $\lambda_m$ is the smallest eigenvalue of $K$. Via adding we get a new matrix $A'$, for which we compute new $K'$ and use the prescribed embedding resulting from $K'$ and not from $K$, when performing $k$-means.  
}

Let us recall the $k$-means quality function which is minimized by $k$-means ($\boldsymbol\mu(C_j)=\boldsymbol\mu_j$ is the gravity center of cluster $C_j$).
 $$Q(\Gamma)=\sum_{j=1}^k \sum_{i\in C_j} ||\mathbf{z}_i-\boldsymbol\mu(C_j)||^2$$ 
 which may be reformulated as 
\begin{equation} \label{eq:Q::kmeans}
Q(\Gamma)
=\sum_{j=1}^k \frac{1}{2n_j} \sum_{i \in C_j} 
\sum_{\ell \in C_j} \|\mathbf{z}_i - \mathbf{z}_\ell\|^2\;  
\end{equation} 
\noindent where $n_j=|C_j|$. 
This implies  
\begin{align*}
Q(\Gamma)=&
\sum_{j=1}^k \frac{1}{2n_j} \sum_{i \in C_j} 
\sum_{\substack{\ell \in C_j\\ \ell \ne i}} 
(1-S_{il})\\
\Bem{
=&
\sum_{j=1}^k \frac{1}{2n_j} \sum_{i \in C_j} 
\sum_{\ell \in C_j; \ell\ne i} 
1-
\sum_{j=1}^k \frac{1}{2n_j} \sum_{i \in C_j} 
\sum_{\ell \in C_j; \ell\ne i} 
S_{i\ell}\\
=&
\sum_{j=1}^k \frac{1}{2n_j} 
n_j (n_j-1)
-
\sum_{j=1}^k \frac{1}{2n_j} \sum_{i \in C_j} 
\sum_{\ell \in C_j; \ell\ne i} 
S_{i\ell}\\}
=&
\frac{n-k}2-\sum_{j=1}^k \frac{1}{2n_j} \sum_{i \in C_j} 
\sum_{\substack{\ell \in C_j\\ \ell \ne i}} \Bem{ {\ell \in C_j; \ell\ne i} }
S_{i\ell}\;.
\end{align*}%
\Bem{
that is 
\begin{equation} \label{eq:Q::kmeansSpectr}
Q(\Gamma)=
\frac{n-k}2-\sum_{j=1}^k \frac{1}{2n_j} \sum_{i \in C_j} 
\sum_{\ell \in C_j; \ell\ne i} 
S_{i\ell}\;
\end{equation}
}%
$n,k$ are independent of clustering. 
We see that $k$-means applied to this $K$-based embedding seeks to find the same clustering as the intention of the $L$-based GCA described in the previous subsection, as both seek to maximize RCut criterion. 
This actually means that if we are able to characterize $K$-based embedding in terms of words, then this description will apply to $L$-based embedding. 

The characterization can be based on the similarity measure which in turn is based on the cosine in term vector space. 
We can assume that the justification of the membership of a document in the cluster is its average similarity to the documents of the same cluster. For cluster $C_j$ and the document $i$ this would mean:
\begin{align*}
    memb(i,C_j)
    =&\frac1{n_j-1}  \sum_{\substack{\ell \in C_j\\ \ell \ne i}}  S_{i\ell}\\
=&1-\frac1{n_j-1}  \sum_{\substack{\ell \in C_j\\ \ell \ne i}}   \|\mathbf{z}_i - \mathbf{z}_\ell\|^2\;.
\end{align*}

\begin{figure}
\begin{center}
\includegraphics[width=0.9\textwidth]{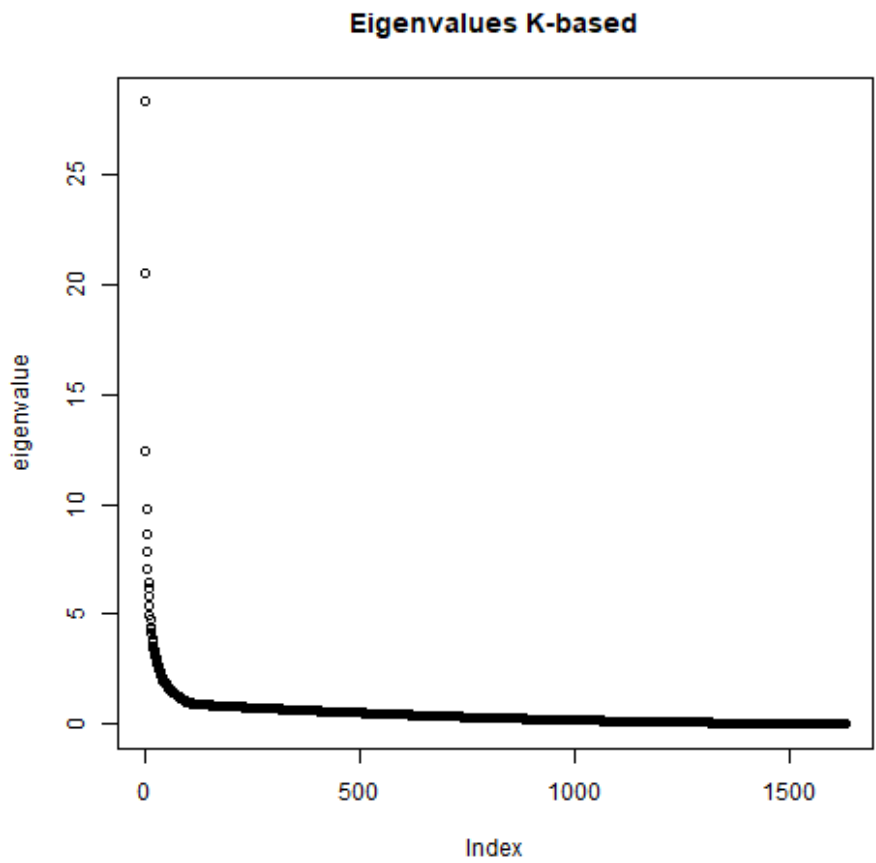} %
 \end{center}
\caption{Distribution of eigenvalues under $K$-based embedding for TWT.4 data 
}\label{fig:eigenvaluesKbased}
\end{figure}

Instead of using all eigenvectors in representing the $K$, the top $m$ eigenvalues and associated eigenvectors can be used to approximate it sufficiently. The reason is the shape of eigenvalue spectrum as visible in Fig. \ref{fig:eigenvaluesKbased} where the leading eigenvalues are much bigger than the other ones.  
In the same spirit, the similarity between cluster center and the concrete element of the cluster may be approximated by top summands in the cosine computation thus pointing to the most important words in the document making it similar to cluster center. 

\begin{figure}
\begin{center}
\includegraphics[width=0.9\textwidth]{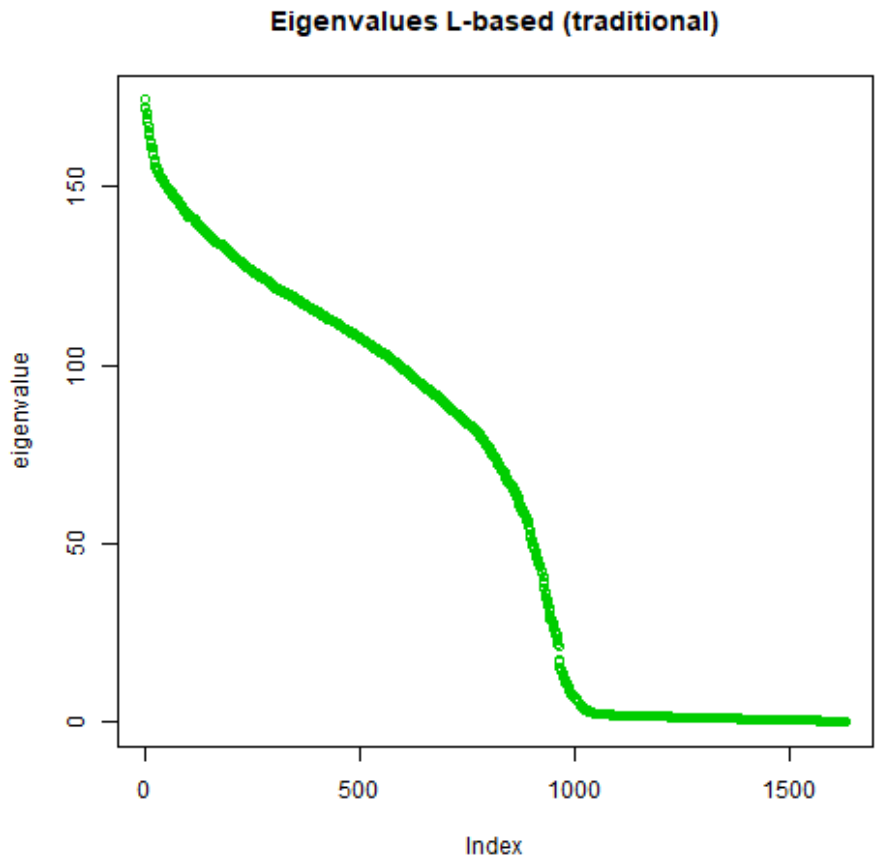} %
 \end{center}
\caption{Distribution of eigenvalues under $L$-based embedding for TWT.4 data 
}\label{fig:eigenvaluesLbased}
\end{figure}

Note that a similar technique (selection of appropriate eigenvalues) was  applied in case of $L$-{embedding}s, see the spectrum of $L$ in Fig.~\ref{fig:eigenvaluesLbased}.

\subsection{Relationship to Document Vector Based Embedding}

Imagine now that $\mathbf{w}_i$ is the embedding of the document $i$ in the original term vector space 
(called afterwards \emph{$W$-embedding})
in which cosine similarities are computed such that $\mathbf{w}_i^T\mathbf{w}_i=1$. Obviously $S_{il}=\mathbf{w}_i^T\mathbf{w}_l$. Furthermore let us denote $\mathbf{w}_{C_j}=\boldsymbol{\mu}_j=\frac1{|C_j|}\sum_{\ell \in C_j} \mathbf{w}_\ell$. Obviously, $\mathbf{w}_{C_j}$ does not need to be a unit vector. 

Let us discuss now the specific distance of a single document from the cluster center:
\[
\begin{split}
 \|\mathbf{w}_i-\boldsymbol{\mu}_j\|^2 
=&(\mathbf{w}_i-\boldsymbol{\mu}_j)^T (\mathbf{w}_i-\boldsymbol{\mu}_j) \\
\Bem{
=&\mathbf{w}_i^T\mathbf{w}_i
-2\mathbf{w}_i^T\boldsymbol{\mu}_j
+\boldsymbol{\mu}_j^T\boldsymbol{\mu}_j
\\
=&\mathbf{w}_i^T\mathbf{w}_i
-2\mathbf{w}_i^T\left(\frac1{|C_j|}\sum_{l \in C_j}\mathbf{w}_l\right)\\
&+\left(\frac1{|C_j|}\sum_{l \in C_j}\mathbf{w}_l\right)^T\left(\frac1{|C_j|}\sum_{l \in C_j}\mathbf{w}_l\right)
\\
=&1
-2\left(\frac1{|C_j|}\sum_{l \in C_j}\mathbf{w}_i^T\mathbf{w}_l\right)\\
&+\left(\frac1{|C_j|}\sum_{l \in C_j}\mathbf{w}_l\right)^T\left(\frac1{|C_j|}\sum_{l \in C_j}\mathbf{w}_l\right)\\
=&1
-2\frac{1}{|C_j|}\left(1+\sum_{\substack{\ell \in C_j\\ \ell \ne i}} S_{i\ell}\right)\\
&+\left(\frac1{|C_j|}\sum_{\ell \in C_j}\mathbf{w}_\ell\right)^T\left(\frac1{|C_j|}\sum_{\ell \in C_j}\mathbf{w}_\ell\right)
\\}
=&
\frac{|C_j|-2}{|C_j|}
-2\frac{1}{|C_j|}\sum_{\substack{\ell \in C_j\\ \ell \ne i}} S_{i\ell}\\
&+\left(\frac1{|C_j|}\sum_{\ell \in C_j}\mathbf{w}_\ell\right)^T\left(\frac1{|C_j|}\sum_{\ell \in C_j}\mathbf{w}_\ell\right)\; .
\end{split}
\]
Note that 
$S_{C_jC_j}=\frac{|C_j|-2}{|C_j|}+\left(\frac1{|C_j|}\sum_{l \in C_j}\mathbf{w}_l\right)^T\left(\frac1{|C_j|}\sum_{l \in C_j}\mathbf{w}_l\right)$
is a constant for cluster $C_j$ and therefore 

\begin{equation}
    \|\mathbf{w}_i-\boldsymbol{\mu}_j\|^2 
    = S_{C_jC_j}-2\frac1{|C_j|}\sum_{\substack{\ell \in C_j\\ \ell \ne i}} S_{i\ell}
\end{equation}
characterizes the closeness to cluster center. 
In this representation on the one hand we can say that the higher the similarities to other documents, the closer the document is to the cluster center. 

How can this property be applied to characterize documents? 
One could think that  most contributive to high average similarity are those words (terms) which have the minimal absolute difference in $\mathbf{w}_i$ to respective coordinate in $\boldsymbol{\mu}_j$. 
However,  missing words may be preferred. 
But this particular embedding has the property that all coordinates are non-negative. Hence the best characterization is by the highest summand when computing the cosine similarity. 
On the other hand cosine summand would be a poor advice if one calculated the difference between a document and other cluster $C_{j'}$. Here the summand maximizing $\|\mathbf{w}_i-\boldsymbol{\mu}_{j'}\|^2 $ would be a good suggestion. 

So we can treat the term vector space embedding vector of a spectral  cluster center as the mean vector of (normalized) cluster components but without normalization. The ``cosine similarity'' to cluster center is not quite cosine, but still we have a legitimate description of the cluster center in terms of words from term vector space. 

In summary, we have pointed out in this section that the traditional $L$-embedding lost the direct relation between {data}point distances and the cosine similarity of documents. This is a serious  disadvantage because $k$-means is applied in GSC clusters based on distances in embedding space, not similarities between documents. 
We have shown that there exists a $K$ embedding having same general goal as $L$-embedding, but with the property that distances in the space are directly translated to similarities so that $k$-means applied in this embedding optimizes on the similarities within a cluster directly. 
And in the third embedding, the term vector space embedding, the similarities can be computed directly as cosine similarity or based on Euclidean distances. This duality allows for precise pointing at sources of similarities of the cluster elements and at sources of dissimilarities in terms of words of the documents.  

In this way the problem of GSC explanation is overcome in that membership reason can be given in terms of sets of decisive words.

\begin{figure}
\begin{center}
\includegraphics[width=0.9\textwidth]{\figaddr{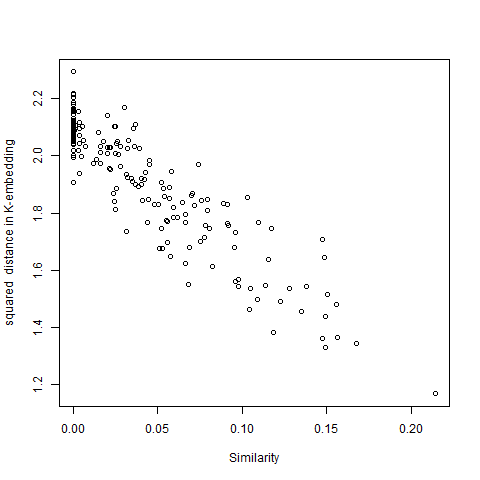}} %
 \end{center}
\caption{Reconstruction of similarity by squared distance under $K$-embedding with the number of coordinates reduced to 400  
}\label{fig:RECeigenvaluesKbased}
\end{figure}

\subsection{A Note on Normalized Laplacian Based Clustering}

The currently more popular $\mathcal{L}$ Laplacian based spectral clustering (see eq. (\ref{eq:normLapDef})) constitutes a bigger challenge for explanation as the translation to cosine similarity is not that straight forward. 

We suggest to use 
the $\mathcal{A}$   matrix of the form:
\begin{equation}
    \mathcal{A}= \mathbf{1}\mathbf{1}^T-I-\mathcal{S}\;.
\end{equation}
\noindent where $\mathcal{S}=D^{-\frac12}SD^{-\frac12}$, with $D$ being defined as previously.  
Let $\mathcal{K}$ be the matrix of the form:
\begin{equation}
    \mathcal{K}=-\frac12(I-\frac1{n}\mathbf{1}\mathbf{1}^T)\mathcal{A}(I-\frac1{n}\mathbf{1}\mathbf{1}^T)\;.
\end{equation}
We proceed with $\mathcal{K}$ as with $K$ matrix. However, the optimization is now not on similarities but on similarities penalized by square roots of volumes of documents, whereas NCut optimizes the quotient of similarity divided by (target) cluster volume. Both are close to one another iff the volumes of documents from the same cluster are equal. Otherwise discrepancies may be expected. If the approximation is good enough, the distance based and cosine product base explanation may be applied as before. 

\section{Experiments}\label{sec:experiments}

Through experiments,  
\begin{itemize}
\item we demonstrate, by inspection of their eigenvalue spectrogram,  that 
$K$ matrix is unrelated to $L$ matrix;
\item 
we demonstrate that $L$-based embedding differs from $K$-based embedding in that $L$-based embedding is poorly related to the similarity measures, while $K$-based embedding is correlated with similarity;
\item 
finally we demonstrate that both are similarly well suited for text clustering in that we show they  they restore groups of tweets sharing same hashtag with similar performance. 
\end{itemize}

\subsection{Data}
\begin{table}
\centering
\begin{tabular}{r|l|c}
\hline
    No.  & hashtag & count \\
    \hline
  0& 90dayfiance & 316\\
	 1& tejran & 345\\
	 2& ukraine & 352\\
	 3& tejasswiprakash & 372\\
	 4& nowplaying & 439\\
	 5& anjisalvacion & 732\\
	 6& puredoctrinesofchrist & 831\\
	 7& 1 & 1105\\
	 8& lolinginlove & 1258\\
	 9& bbnaija & 1405\\
  \hline
\end{tabular}
\caption{TWT.10 data set - hashtags and cardinalities of the set of related tweets  used in the experiments}\label{tab:twt10set}
\end{table}

In the experiments, we use three sets of data:
\begin{itemize}
    \item A synthetic dataset BLK of 2000 ``product descriptions'' divided into 4 classes; the dataset was generated by a random generator providing with random descriptive texts, but characterized by a clear block-structure relationships within the classes (generator in the \texttt{Supplementary File}).
    \item The set TWT.4, being a collection of tweets related to hashtags 
    \#anjisalvacion, 
  \#lolinginlove, 
  \#nowplaying and   
  \#puredoctrinesofchrist from TWT.10 dataset. 
  \item The set TWT.10, being a collection of tweets related to hashtags listed in Table~\ref{tab:twt10set} (available in \texttt{Supplementary File})\footnote{submitted together with this paper, to be made public later.}.  
\end{itemize}

\begin{table}
     \centering
     \begin{tabular}{|r|rrrr|}
\hline 
            & \multicolumn{4}{c|}{clsLbased}  \\
            \cline{2-5}
clsKbased&    1   &2   &3   &4\\
      \hline
      1&   0&  86& 400&   0\\
      2&   500&   0&   0& 0\\
      3&   0&   0&   0& 500\\
      4&   0& 513&   0&   1\\
          \hline
     \end{tabular}
     \caption{Confusion matrix;
     clsLbased  - clusters generated from $L$-embedding, 
     clsKbased - clusters generated from $K$-embedding, 
     number of elements in correct clusters: 1913, 
incorrectly clustered:  87, 
errors:  4.35 \% .
Parameter:  $r=number\ of\ clusters+1$. 
}
     \label{tab:confKL}
 \end{table}

\subsection{Differences between $L$- and $K$-embeddings}
First, we have computed the specrograms of the $K$-matrix and $L$-matrix for the TWT.4 dataset. 
They are shown in Figures~\ref{fig:eigenvaluesKbased} and \ref{fig:eigenvaluesLbased} respectively. 
One can see that the shapes of these spectrograms differ strongly so that it cannot be claimed that $K$-embedding based clustering and $L$-embedding based clustering rely on related mathematical concepts. 

\subsection{Relationship of $K$-embedding clustering and $L$-embedding clustering}

We performed experiments using our synthetic data  BLK tat that  creates block data, friendly for GSC methodology. 

We clustered the data using the traditional GSC clustering method ($L$-embedding) and using our one ($K$-embedding). The results comparing the clusterings produced by each of them  are presented in Table~\ref{tab:confKL}. Results of $K$-embedding are taken as "ground truth".  We see that the clustering largely agree. 
We compared also both the "real" groups in tables \ref{tab:confTrueL} and  \ref{tab:confTrueK}. We see that $K$-embedding based approach gets closer to real groups than $L$-based approach.

 \begin{table}
     \centering
     \begin{tabular}{|r|rrrr|}
\hline 
            &  \multicolumn{4}{c|}{clsKbased}    \\
             \cline{2-5}
clsTrue&    1   &2   &3   &4\\
      \hline
      1&   0& 500&   0&   0\\
      2&   0&   0& 500&   0\\
      3&   1&   0&   0& 499\\
      4& 485&   0&   0&  15\\
      \hline
\end{tabular}
\caption{
Confusion matrix;
     clsKbased - clusters generated from $K$-embedding, 
     clsTrue  - the true clusters, number of elements in correct clusters: 1984, incorrectly clustered:  16, error:  0.8 \% 
}
     \label{tab:confTrueK}
 \end{table}

 \begin{table}
     \centering
     \begin{tabular}{|r|rrrr|}
\hline 
            &  \multicolumn{4}{c|}{clsLbased}  \\
             \cline{2-5}
clsTrue&    1   &2   &3   &4\\
      \hline
      1& 500&   0&   0&   0\\
      2&   0&   0&   0& 500\\
      3&   0& 499&   0&   1\\
      4&   0& 100& 400&   0\\
      \hline
\end{tabular}
\caption{
 Confusion matrix;
     clsLbased  - clusters generated from $L$-embedding, 
     clsTrue  - the true clusters,  
number of elements in correct clusters 1899 
incorrectly clustered:  101 
= errors:  5.05 %
}
     \label{tab:confTrueL}
 \end{table}

We performed also experiments using real world data set TWT.4 using both mentioned methods of clustering. 
 The results,  comparing the clusterings produced by each of them against the ground truth, being the hashtag groups, are presented in Tables~\ref{tab:confTrueKrealdata} and~\ref{tab:confTrueLrealdata}.   We see that $K$-embedding based approach gets closer to real groups than $L$-based approach. 

In summary we can say that our method gets more closely to the real clustering than the conventional GSC. 

 \begin{table}
     \centering
     \begin{tabular}{|r|rrrr|}
\hline 
            &  \multicolumn{4}{c|}{clsLbased}  \\
             \cline{2-5}
clsKbased&    1   &2   &3   &4\\
      \hline
      1& 202 &  0 &  0  & 0\\
      2& 660 &  0 &  0  & 0\\
      3& 740 &  1 &  1  & 1\\
      4&  25 &  0 &  0 &  0\\
      \hline
\end{tabular}
\caption{
 Confusion matrix;
     clsLbased  - clusters generated from $L$-embedding, 
     clsKbased  - clusters generated from $K$-embedding, 
     number of elements in correct clusters: 743, 
incorrectly clustered:  887, errors:  54.4 \% 
}
     \label{tab:confKLrealdata}
 \end{table}

\begin{table}
    \centering
    \begin{tabular}{|r|rrrr|}
        \hline 
                    &  \multicolumn{4}{c|}{clsKbased}  \\
        \cline{2-5}
        clsTrue                 & 1  & 2  & 3  & 4  \\
        \hline
        \#anjisalvacion         & 370&   0&   0&  10\\
        \#lolinginlove          &   0&   0&   0& 614\\
        \#nowplaying            &  85&  27&   1&  96\\
        \#puredoctrinesofchrist & 207&   0& 203&  17\\
        \hline
    \end{tabular}
\caption{
Confusion matrix;
clsKbased - clusters generated from $K$-embedding, 
clsTrue  - the true clusters,  
number of elements in correct clusters: 1214 
incorrectly clustered:  416, 
errors:  25.52147 \% 
}
\label{tab:confTrueKrealdata}
\end{table}

 \begin{table}
     \centering
     \begin{tabular}{|r|rrrr|}
\hline 
            &  \multicolumn{4}{c|}{clsLbased}  \\
             \cline{2-5}
clsTrue&    1   &2   &3   &4\\
      \hline
  \#anjisalvacion           &0   &0   &0 &380\\
  \#lolinginlove            &1   &3   &2 &608\\
  \#nowplaying              &0   &0   &0 &209\\
  \#puredoctrinesofchrist   &0   &0   &0 &427\\
      \hline
\end{tabular}
\caption{
 Confusion matrix;
     clsLbased - clusters generated from $L$-embedding, 
     clsTrue  - the true clusters,  
number of elements in correct clusters 614 
incorrectly clustered:  1016 
errors:  62.33129 \% 
}
     \label{tab:confTrueLrealdata}
 \end{table}

\Bem{
This experiment was performed on   best predicted hashtags which are are: \#bbnaija, 
\#lolinginlove, 
\#anjisalvacion and  
\#puredoctrinesofchrist
}

\subsection{Discrepancies Between Embeddings and Similarities}

\begin{figure}
\begin{center}
\includegraphics[width=0.3\textwidth]{\figaddr{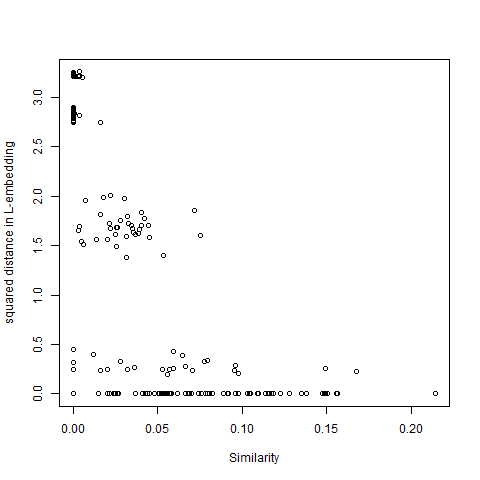}} %
 \end{center}
\caption{Reconstruction of similarity by squared distance under $K$-embedding with the number of coordinates reduced to 5 (as required by GSC)  
}\label{fig:RECeigenvaluesLbased}
\end{figure}

We have also investigated the relationships between the $L$-embedding and the similarities and the $K$-embeddings and similarities for the BLK dataset. 
We randomly selected 80 ``documents'' and drew in Figures~\ref{fig:RECeigenvaluesKbased} and \ref{fig:RECeigenvaluesLbased} plots of the distances in the embeddings and the document similarities. 
\begin{table}
\centering
\begin{tabular}{|r|r|r|}
\hline
       r&         errorK&        errorL\\
       \hline
  1000& 0.001713712& 0.03710026\\
   500& 0.003453155& 0.03710026\\
   250& 0.004585440& 0.03710026\\
   125& 0.005164994& 0.03710026\\
    62& 0.005363871& 0.03710026\\
    31&0.005285830 &0.03710026\\
    16& 0.005282382& 0.03710026\\
     8& 0.005252960& 0.03710026\\
    4 &0.005268428& 0.03710026\\
\hline
\end{tabular}
\caption{Errors in reconstructing the K (errorK) and L (errorL) matrix resp. by the subsets of eigenvectors and eigenvalues of various cardinalities (r). They indicate that the K reconstruction is better. 
}\label{tab:reconstructionerrors}
\end{table}
One sees that the distances in the $K$-embedding are more closely related to similarities than those of $L$-embeddings. This confirms  that we cannot explain the document membership in a cluster based on $L$-embeddings, while $K$-embedding justifies such an interpretation of clusters. 

Note that $L$-embedding has the good side that one allows to perform clustering in low dimensions. This may be of course the reason why the relationship of $L$-embedding to similarities is so poor though the $L$ matrix itself contains these similarities. 
In Table~\ref{tab:reconstructionerrors} we compared the $K$ matrix and $L$-matrix reconstruction mean absolute errors when using different number of eigenvectors (top eigenvectors for $K$, and low eigenvectors for $L$). 
As one might have expected, for various numbers of eigenvectors, the reconstruction in $K$ is better than in $L$. 
\begin{table}
\centering
\begin{tabular}{lrr}
\hline 
Score & nearest\_neighbors & precomputed\\
  \hline  
 adjusted mutual info score: & \textbf{0.486238}  &	0.400122\\ 
adjusted rand score:     & \textbf{ 0.308131}  	&0.139347\\
completeness score:     &   \textbf{ 0.394151}	&0.268025\\
fowlkes mallows score     & \textbf{ 0.492279}	&0.433780\\
homogeneity score:      &   0.640522	&\textbf{0.808333}\\
mutual info score:        & \textbf{ 0.849066}	&0.577371\\
normalized mutual info score: 	&\textbf{0.488004}&0.402568\\
rand score:        &      \textbf{ 0.733106} 	&0.481941\\
v measure score:       &  \textbf{ 0.488004}   	&0.402568\\
F-score: &  \textbf{0.1105}  &0.0413    \\
\hline
F-score average: & \textbf{  0.0596} &  0.0291 \\
\hline 
\end{tabular}
\caption{$L$-based spectral clustering scores under diverse settings of affinity parameter (column names)}
\label{tab:twt10spectral}
\end{table}

\subsection{Clustering Performance for 10 hashtags}

The experiments were performed on TWT.10 Twitter data for a selected set of 10 hashtags that had to appear only once in the tweet. 
The hashtags are listed in Table~\ref{tab:twt10set}. 
The reason for choice of such tweets was to have a human induced reference set on which the quality of clustering was evaluated. 

The clustering experiments were performed with popular Python libraries: 
numpy \cite{NumPy:2020}, scipy \cite{SciPy:2020}, scikit-learn \cite{sklearnAPI:2013} and 
soyclustering \cite{soyclustering:020} which is an implementation of spherical $k$-means \cite{SKmeans:2020:113288}.
In particular, we used 

\begin{enumerate}
    \item \texttt{SpectralClustering} class from scikit-learn with two distinct  settings of the \texttt{affinity} parameter: \texttt{precomputed} (affinity from similarity matrix) and \texttt{nearest\_neighbors} (affinity from graph of nearest neighbors) - as a representative of the $L$-embedding based clustering (see Table~\ref{tab:twt10spectral}), and 
    
    \item \texttt{SphericalKMeans} class from soyclustering with the following combinations of (\texttt{init}, \texttt{sparsity}) parameter pairs (short names given for reference in Table~\ref{tab:twt10Kembed.r10} and following):
           "sc.n": ('similar\_cut', None),
           "sc.sc": ('similar\_cut', 'sculley'),
           "sc.md": ('similar\_cut', 'minimum\_df'),
           "k++.n": ('k-means++', None),
           "k++.sc": ('k-means++', 'sculley'),
           "k++.md": ('k-means++', 'minimum\_df'), and 
    \item  $K$-embedding clustering (our implementation, exploiting spherical $k$-means).
    The following numbers of eigenvectors were tried: $r=10$ (number of hashtags), $r=20$, $r=3577$ (half of tweet count) and $r=7155$ (tweet count). 
    See Tables~\ref{tab:twt10Kembed.r10}, \ref{tab:twt10Kembed.r20}, \ref{tab:twt10Kembed.r3577} and \ref{tab:twt10Kembed.r7155}. 
\end{enumerate}

\noindent 

\Bem{
For each algorithm and each parameter configuration, 10 runs were performed and the best performing run as presented in terms of F1 value. 

\begin{tabular}{lr}
\hline
adjusted mutual info score:  & 	0.486238\\
adjusted rand score:   &       	0.308131\\
completeness score:      &     	0.394151\\
fowlkes mallows score     &    	0.492279\\
homogeneity score:      &      	0.640522\\
mutual info score:      &      	0.849066\\
normalized mutual info score: 	& 0.488004\\
rand score:      &             	0.733106\\
v measure score:   &           	0.488004\\
F-score: & 0.1105\\
\hline
\end{tabular}
The spectral clustering with affinity=precomputed yielded:
}

\begin{table}
\centering{\tiny
\begin{tabular}{lrrr}
\hline
Score & sc.n  &  sc.sc & sc.md   \\
  \hline  
adjusted mutual info score      & 0.442456   & \textbf{ 0.481259}   & 0.450716   \\
adjusted rand score             & 0.370633   & \textbf{ 0.409385}  & 0.389930   \\
completeness score              & 0.452317   & \textbf{ 0.482101}  & 0.448153   \\
fowlkes mallows score           & 0.454743   & \textbf{ 0.485733}  & 0.467060   \\
homogeneity score               & 0.435729   & \textbf{ 0.483157}  & 0.456226   \\
mutual info score               & 0.938634   & \textbf{ 1.040801}  & 0.982786   \\
normalized mutual info score    & 0.443868   & \textbf{ 0.482629}  & 0.452153   \\
rand score                      & 0.854369   &  \textbf{0.867059}  & 0.865477   \\
v measure score                 & 0.443868   & \textbf{ 0.482629}  & 0.452153   \\
F-score                         & 0.185137   & \textbf{ 0.235635}  & 0.180591   \\
 \hline 
F-score average:                         &  0.0881  &  \textbf{0.1077}  &   0.1070   \\
\hline
\hline
Score &  k++.n&  k++.sc  & k++.md \\
  \hline  
adjusted mutual info score      &  0.377781  &  0.430736   &   0.412699   \\
adjusted rand score             &  0.352018  &  0.391106   &   0.377490   \\
completeness score              &  0.375782  &  0.428704   &   0.414931   \\
fowlkes mallows score           &  0.433405  &  0.467857   &   0.456492   \\
homogeneity score               &  0.383106  &  0.435809   &   0.413615   \\
mutual info score               &  0.825274  &  0.938805   &   0.890996   \\
normalized mutual info score    &  0.379409  &  0.432227   &   0.414272   \\
rand score                      &  0.857969  &  0.866138   &   0.862212   \\
v measure score                 &  0.379409  &  0.432227   &   0.414272   \\
F-score                         &  0.162810  &  0.166237   &   0.198266   \\
 \hline 
F-score average:                         &  0.0937  &  0.0765   &   0.0974   \\
  \hline  
\end{tabular}}
\caption{Spherical $k$-means,  achieved 
scores under diverse settings of the algorithm;
the highest values in 10 runs.}
\label{tab:twt10skameans}
\end{table}

As visible from Tables~\ref{tab:twt10spectral} and \ref{tab:twt10skameans},   
the spherical $k$-means worst F-score (0.16281 for "k++.n") is superior to the best spectral clustering score (0.1105 for nearest\_neighbors). 
$K$-embedding based clustering (tables \ref{tab:twt10Kembed.r10}--\ref{tab:twt10Kembed.r7155}) 
best F-score achieved was 0.2144 for "sc.sc" and r=7155.


This experiment demonstrates that $K$-embedding clustering can approximate a real world data clustering at a level at least comparable with $L$-based clustering, so that the clustering explanation bridge $L$-embedding - $K$-embedding - term vector space embedding appears to be justified. 

\Bem{
\begin{table}
\centering
\begin{tabular}{|l|r|r|r|r|}
\hline
 & r=10  & r=20 & r=3577 &r=7155\\
config & F1 & F1 & F1 & F1\\
\hline
\hline
  sc.n  &  0.1831 &  0.1642 & \textbf{ 0.1923}&  0.2285\\
  sc.sc & \textbf{ 0.2174}& \textbf{ 0.1767}&  0.1312&  0.1553\\
  sc.md & 0.1785& 0.1633& 0.1712& 0.1891\\
  k++.n & 0.1268& 0.1196& 0.1208& \textbf{ 0.3150}\\
  k++.sc & 0.1266& 0.1542& 0.1841& 0.1969\\
  k++.md &  0.1203&  0.1012&  0.0941&  0.2759\\
  \hline
\end{tabular}
\caption{$K$-based clustering,  
F-Score under diverse settings of spherical $k$-means algorithm (rows) and the number of dimensions used (columns); the highest values in 10 runs. 
}\label{tab:twt10Kembed.r10}
\end{table}
} 

\begin{table}
\centering{\tiny
\begin{tabular}{lrrr}
\hline
Score &  sc.n  &  sc.sc  & sc.md \\
  \hline  
  adjusted mutual info score  &  0.259935  &  0.273119  & 0.259381 \\
  adjusted rand score  & \textbf{ 0.261349} & 0.241035  & 0.236007 \\
  completeness score  &  0.260018  & 0.268650  & 0.256508 \\
  fowlkes mallows score  & \textbf{ 0.362643} & 0.352751  & 0.346052 \\
  homogeneity score  &  0.263817  & 0.281839  & 0.266431 \\
  mutual info score  &  0.560122  & 0.578717  & 0.552560 \\
  normalized mutual info score  &  0.261904  & 0.275087  & 0.261375 \\
  rand score  & \textbf{ 0.825399} & 0.810171  & 0.812104 \\
  v measure score  &  0.261904  & 0.275087  & 0.261375 \\
  F-score  &  0.1635  & 0.1335  & 0.1348 \\
\hline
  F-score average: & 0.1005 & 0.0904 & 0.0757 \\
 \hline
\hline
Score &  k++.n  &  k++.sc  & k++.md \\
  \hline  
  adjusted mutual info score  & \textbf{ 0.282358} & 0.255432  & 0.259865 \\
  adjusted rand score  &  0.252918  & 0.238203  & 0.241569 \\
  completeness score  & \textbf{  0.278569} & 0.253377  & 0.257751 \\
  fowlkes mallows score  &  0.360486  & 0.347467  & 0.349507 \\
  homogeneity score  & \textbf{ 0.290259} & 0.261615  & 0.266085 \\
  mutual info score  & \textbf{  0.600085} & 0.545816  & 0.555239 \\
  normalized mutual info score  & \textbf{  0.284294} & 0.257430  & 0.261852 \\
  rand score  &  0.816440  & 0.813288  & 0.815278 \\
  v measure score  & \textbf{ 0.284294} & 0.257430  & 0.261852 \\
  F-score  & \textbf{ 0.2119} & 0.1624  & 0.1396 \\
  \hline
  F-score average: & 0.0850 & \textbf{ 0.1048} & 0.0861 \\
  \hline  
\end{tabular}}
\caption{ 
$K$-based clustering,  
scores (rows) under diverse settings of spherical $k$-means algorithm (columns), when   the number of dimensions used (columns) r=10; the highest values in 10 runs. 
}\label{tab:twt10Kembed.r10}
\end{table}

\begin{table}
\centering{\tiny
\begin{tabular}{lrrr}
\hline
Score &  sc.n  &  sc.sc  & sc.md \\
  \hline  
  adjusted mutual info score  & \textbf{ 0.299142} & 0.266797  & 0.263523 \\
  adjusted rand score  & \textbf{ 0.266636}  & 0.262783  & 0.242990 \\
  completeness score  & \textbf{ 0.296987}  & 0.265751  & 0.259459 \\
  fowlkes mallows score  & \textbf{  0.369347} & 0.364581  & 0.352675 \\
  homogeneity score  & \textbf{ 0.305162}  & 0.271831  & 0.271862 \\
  mutual info score  & \textbf{ 0.639759} & 0.572471  & 0.558917 \\
  normalized mutual info score  & \textbf{ 0.301019} & 0.268757  & 0.265516 \\
  rand score  & \textbf{ 0.823745} & 0.824777  & 0.813006 \\
  v measure score  & \textbf{ 0.301019} & 0.268757  & 0.265516 \\
  F-score  & 0.1509   & \textbf{  0.1996} & 0.1925 \\
\hline 
  F-score average:  &  0.0808  & \textbf{  0.1146} & 0.0969 \\
\hline
\hline
Score &  k++.n  &  k++.sc  & k++.md \\
  \hline  
  adjusted mutual info score  &  0.276224  & 0.261085  & 0.274141 \\
  adjusted rand score  &  0.243241  & 0.256841  & 0.258908 \\
  completeness score  &  0.273747  & 0.262463  & 0.274735 \\
  fowlkes mallows score  &  0.351636  & 0.357758  & 0.360017 \\
  homogeneity score  &  0.282733  & 0.263620  & 0.277413 \\
  mutual info score  &  0.589696  & 0.565390  & 0.591826 \\
  normalized mutual info score  &  0.278168  & 0.263040  & 0.276068 \\
  rand score  &  0.814755  & 0.825707  & 0.825537 \\
  v measure score  &  0.278168  & 0.263040  & 0.276068 \\
  F-score  &  0.1458  & 0.1400  & 0.1061 \\
\hline 
  F-score average:  &  0.0860  & 0.0844  & 0.0731 \\
  \hline  
\end{tabular}}
\caption{ $K$-based clustering,  
scores (rows) under diverse settings of spherical $k$-means algorithm (columns), when   the number of dimensions used (columns) r=20; the highest values in 10 runs. 
}\label{tab:twt10Kembed.r20}
\end{table}

\begin{table}
\centering{\tiny
\begin{tabular}{lrrr}
\hline
Score &  sc.n  &  sc.sc  & sc.md \\
  \hline  
  adjusted mutual info score  & \textbf{  0.280581} & 0.251402  & 0.275928 \\
  adjusted rand score  & \textbf{  0.265739}  & 0.237246  & 0.270445 \\
  completeness score  & \textbf{ 0.278901}  & 0.249544  & 0.277554 \\
  fowlkes mallows score  & \textbf{ 0.368818}  & 0.346103  & 0.367983 \\
  homogeneity score  & \textbf{ 0.286205}  & 0.257399  & 0.278130 \\
  mutual info score  & \textbf{ 0.600799}  & 0.537559  & 0.597898 \\
  normalized mutual info score  & \textbf{ 0.282506}  & 0.253410  & 0.277842 \\
  rand score  &  0.823199  & 0.813786  & \textbf{ 0.831067} \\
  v measure score  & \textbf{ 0.282506}  & 0.253410  & 0.277842 \\
  F-score  &  0.1515  & \textbf{ 0.1666} & 0.1146 \\
  \hline
  F-score average:  &  \textbf{0.1011}  &  0.0845  & 0.0677 \\
  \hline
\hline
Score &  k++.n  &  k++.sc  & k++.md \\
  \hline  
  adjusted mutual info score  &  0.273632  & 0.278102  & 0.263998 \\
  adjusted rand score  &  0.260796  & 0.245184  & 0.260272 \\
  completeness score  &  0.273543  & 0.275847  & 0.265390 \\
  fowlkes mallows score  &  0.362615  & 0.352776  & 0.360405 \\
  homogeneity score  &  0.277616  & 0.284359  & 0.266504 \\
  mutual info score  &  0.589257  & 0.594221  & 0.571695 \\
  normalized mutual info score  &  0.275565  & 0.280039  & 0.265946 \\
  rand score  &  0.824647  & 0.815950  & 0.826961 \\
  v measure score  &  0.275565  & 0.280039  & 0.265946 \\
  F-score  &  0.1410  & 0.1302  & 0.1129 \\
  \hline
  F-score average:  &  0.0885  & 0.0666  & 0.0682 \\
  \hline  
\end{tabular}}
\caption{
$K$-based clustering,  
scores (rows) under diverse settings of spherical $k$-means algorithm (columns), when   the number of dimensions used (columns) r=3577; the highest values in 10 runs. 
}\label{tab:twt10Kembed.r3577}
\end{table}

Finally, we have checked which clustering results are closer to those of $K$-based clustering - it turned out that spherical clustering is closer than spectral one, see Table~\ref{tab:towhatisKbasedcloser}. 

\subsection{A Discussion}
The selected real world dataset TWT.10 was in general not friendly for spectral clustering methods. Nonetheless it points out that our $K$-embedding can be a candidate for substitution of $L$-embedding for such cases. On the other hand, the GSC friendly (artificial) dataset supports our claim that the explanation path that we have proposed is justified. 

\section{Conclusions}\label{sec:conclusions}

We have constructed a theoretical bridge linking the clusters resulting from Graph Spectral Clustering and the actual document content, given that similarities between documents are computed as cosine measures in tf or tfidf representation. This link enables to justify textually  the   cluster membership of a document derived from cosine similarity and at the same time to justify textually non-membership in other clusters via distance computation in document vector embedding space. 

This result is based on a comparative study of three different {embedding}s of documents: one in the term vector space, one in the spectral clustering space and one based on kernel approach. 
The kernel based approach shares with the term vector space approach the reproduction of cosine similarity while performing traditional $k$-means clustering, but has much lower dimensionality. On the other hand, both kernel-based approach and spectral analysis based approach use the traditional (distance-based)  $k$-means at their heart, and approximate the same target function. 

An important question for future research is the issue why some number of clusters have been chosen. 
There have been some research on automated selection of the number of clusters in general \cite{Fang:2012} and in spectral clustering domain \cite{Alshammari:2019}, but the results for real data that were available to us were not satisfactory so that this topic should be considered as future research area,

\begin{table}
\centering{\tiny
\begin{tabular}{lrrr}
\hline
Score &  sc.n  &  sc.sc  & sc.md \\
  \hline  
  adjusted mutual info score  &  0.351894  & 0.385909  & 0.284980 \\
  adjusted rand score  &  0.331422  & 0.357056  & 0.285706 \\
  completeness score  &  0.356581  & 0.391112  & 0.288709 \\
  fowlkes mallows score  &  0.417609  & 0.439716  & 0.378092 \\
  homogeneity score  &  0.350650  & 0.383984  & 0.285026 \\
  mutual info score  &  0.768135  & 0.842521  & 0.621927 \\
  normalized mutual info score  &  0.353591  & 0.387516  & 0.286856 \\
  rand score  &  0.849867  & 0.855971  & 0.839133 \\
  v measure score  &  0.353591  & 0.387516  & 0.286856 \\
  F-score  &  0.1051  & \textbf{ 0.2144} & 0.1317 \\
  \hline 
  F-score average: &  0.0662  & 0.0930  & 0.0757 \\
\hline
\hline
Score &  k++.n  &  k++.sc  & k++.md \\
  \hline  
  adjusted mutual info score  & \textbf{ 0.401542} & 0.360389  & 0.356130 \\
  adjusted rand score  &  \textbf{0.377350}  & 0.340112  & 0.335164 \\
  completeness score  & \textbf{ 0.408082} & 0.364872  & 0.360456 \\
  fowlkes mallows score  & \textbf{ 0.456293} & 0.425090  & 0.420925 \\
  homogeneity score  &  0.398242  & \textbf{0.359300}  & 0.355218 \\
  mutual info score  &  \textbf{0.879077}  & 0.785996  & 0.776483 \\
  normalized mutual info score  & \textbf{ 0.403102} & 0.362065  & 0.357818 \\
  rand score  & \textbf{ 0.862309} & 0.851955  & 0.850621 \\
  v measure score  & \textbf{ 0.403102} & 0.362065  & 0.357818 \\
  F-score  &  0.1864  & 0.2015  & 0.1642 \\
  \hline 
  F-score average: &  0.0913  & 0.0933  & \textbf{0.1000} \\
  \hline  
\end{tabular}}
\caption{ $K$-based clustering,  
scores (rows) under diverse settings of spherical $k$-means algorithm (columns), when   the number of dimensions used (columns) r=7155; the highest values in 10 runs. 
}\label{tab:twt10Kembed.r7155}
\end{table}

\begin{table}
\begin{tabular}{lrrr}
\hline
& \multicolumn{2}{c}{F1 for spectral} & F1 for spherical  \\
            \cline{2-3}
config &  precomp & nn & \\
\hline
  sc.n   &  0.0604 & 0.1129 & \textbf{0.1820} \\
  sc.sc   &  0.1101 & 0.1343 & \textbf{0.1422}\\
  sc.md & 0.1315 & 0.1469 & \textbf{0.1627}\\
  k++.n & 0.0940 & 0.1042 & \textbf{0.1358}\\
  k++.sc &  0.0612 & 0.1359 & \textbf{0.2608}\\
  k++.md &  0.1034 & 0.1754 & \textbf{0.1817}\\
  \hline
\end{tabular}
\caption{Best 
F-score for predicting spectral clustering and spherical clustering by $K$-based clustering for r= 3754.}
\label{tab:towhatisKbasedcloser}
\end{table}

\bibliographystyle{plain}


\end{document}